\documentclass[11pt]{article}

\usepackage[preprint]{acl}

\usepackage{times}
\usepackage{latexsym}
\usepackage[T1]{fontenc}
\usepackage[utf8]{inputenc}
\usepackage{microtype}
\usepackage{inconsolata}
\usepackage{graphicx}
\usepackage{booktabs}
\usepackage{amsmath}
\usepackage{amsfonts}
\usepackage{enumitem}

\usepackage{tabularx}
\usepackage{booktabs}

\title{SG-Blend: Learning an Interpolation Between Improved Swish and GELU for Robust Neural Representations}

\author{
  Gaurav Sarkar \\
  \texttt{\small gaurav.sarkar@research.iiit.ac.in}
  \And
  Syed Affan Daimi \\
  \texttt{\small syed.daimi@research.iiit.ac.in}
  \AND
  Jay Gala \\
  \texttt{\small t-jaygala@microsoft.com}
  \And
  Subarna Tripathi \\
  \texttt{\small subarna.tripathi@intel.com}
}

\begin{document}
\maketitle

\begin{abstract}

Prevailing activation functions such as Swish and GELU tend toward
domain-specific optima, Swish was discovered via neural architecture
search on vision benchmarks, while GELU dominates transformer-based
language models, and neither offers any mechanism to adapt its
gating shape to individual layers. This rigidity is especially
consequential in transformer FFN blocks, where LayerNorm, unlike
BatchNorm, does not suppress the gradient pathologies that activation
choice induces across depth.
We propose \textbf{SG-Blend}, a per-layer adaptive activation that
combines \emph{SSwish}, a bias-corrected, parametric Swish variant
we also introduce, with learnable sharpness $\beta$ and
zero-centering bias $\gamma$, with GELU through a per-layer blend
coefficient $\alpha$, letting each layer locate its own optimum along
the SSwish--GELU continuum at a cost of only three additional scalars
per FFN block, with $\beta$ initialized to $1.0$ and learned freely
via backpropagation.
On BERT-style IMDB classification 
(5~seeds), it matches peak accuracy ($81.31\%$) while reducing 
seed-to-seed variance by $42\%$ relative to GELU. Furthermore, it generalizes to autoregressive 
pretraining, achieving the lowest validation perplexity ($49.10$) on 
WikiText-103 among all baselines. Crucially, ablations confirm the 
interpolation structure itself drives these gains, delivering reliable, 
top-tier performance. Beyond natural language processing, we demonstrate that SG-Blend generalizes robustly to a wider variety of tasks, extending its efficacy to computer vision and other diverse domains.

\end{abstract}

\section{Introduction}
\label{sec:intro}

Across diverse domains from natural language processing to computer vision, deep architectures are constrained by their reliance on fixed-shape activation functions. Although Swish~\citep{ramachandran-2018} and GELU~\citep{GELU_Hendrycks2016GaussianEL} have emerged as the standard choices across modern networks, they apply the exact same nonlinear gating across all layers. This structural rigidity forces every layer to use an identical activation profile regardless of its depth or domain, leading to inconsistent gradient flow and high sensitivity to random initialization. Consequently, models suffer from severe seed-to-seed performance variance.
In our BERT IMDB study (Section~\ref{sec:bert_imdb}), GELU's
worst-case trails its best by $1.42$\,pp, enough to reverse a
head-to-head comparison.

The instability traces to a design mismatch.
Swish was discovered via neural architecture search on vision benchmarks;
GELU was motivated as the expectation of a stochastic regularizer, specifically,
the expected transformation of Adaptive Dropout applied to the
input~\citep{GELU_Hendrycks2016GaussianEL}.
Neither was designed for the LayerNorm setting that governs modern
transformer FFN stacks: unlike BatchNorm, LayerNorm does not normalize
pre-activations across the batch dimension, so the gradient-stabilizing
effect that co-evolved with these functions in CNN pipelines is absent.
More concretely, both functions apply \emph{identical} nonlinear gating
at every one of a BERT-large model's 24 FFN layers, regardless of
depth, role, or activation distribution.




We propose \textbf{SG-Blend}, which lets each layer learn its own balance
between two well-understood activations: a bias-corrected Swish variant
(SSwish) and GELU.
The blend coefficient $\alpha$ is learnable per layer, co-optimized with
a sharpness parameter $\beta$ and a bias-shift parameter $\gamma$.
Three scalars per FFN block trained end-to-end reduce BERT IMDB
seed-to-seed variance by $42\%$ relative to GELU at matched mean accuracy.

\paragraph{Relation to GELU.}
It is well-known that $\mathrm{GELU}(x) \approx x\,\sigma(1.702\,x)$,
i.e.\ GELU can be viewed as a special case of Swish with a fixed
$\beta{=}1.702$.%
\footnote{GELU is also \emph{non-monotonic}: it has a local minimum at
$x \approx -0.75$ with value ${\approx}{-}0.17$.}
Under this lens, SG-Blend learns the interpolation
$x[\alpha\sigma(\beta x) + (1{-}\alpha)\sigma(1.702\,x)] - \alpha\gamma$,
where $\beta$ and $\alpha$ are learnable.
We do \emph{not} collapse the two paths into one sigmoid because the separate
parameterization makes $\alpha$ directly interpretable as a per-layer
preference score between two well-understood activation families, enabling
the learned-parameter analysis in Section~\ref{sec:param_analysis}.
\paragraph{Sharpness parameter $\beta$.}
$\beta$ is initialized to $1.0$ (matching the GELU-equivalent curvature)
and learned freely via backpropagation, projected to $[0.1, 10]$ after
each gradient step.
Our learned-parameter analysis (Section~\ref{sec:param_analysis}) shows
that $\beta$ self-organizes to near $1.0$ across all layers and seeds,
confirming that gradient learning naturally discovers GELU-like curvature
as the optimum for transformer FFN blocks.

\begin{figure}[t]
  \centering
  \includegraphics[width=\columnwidth]{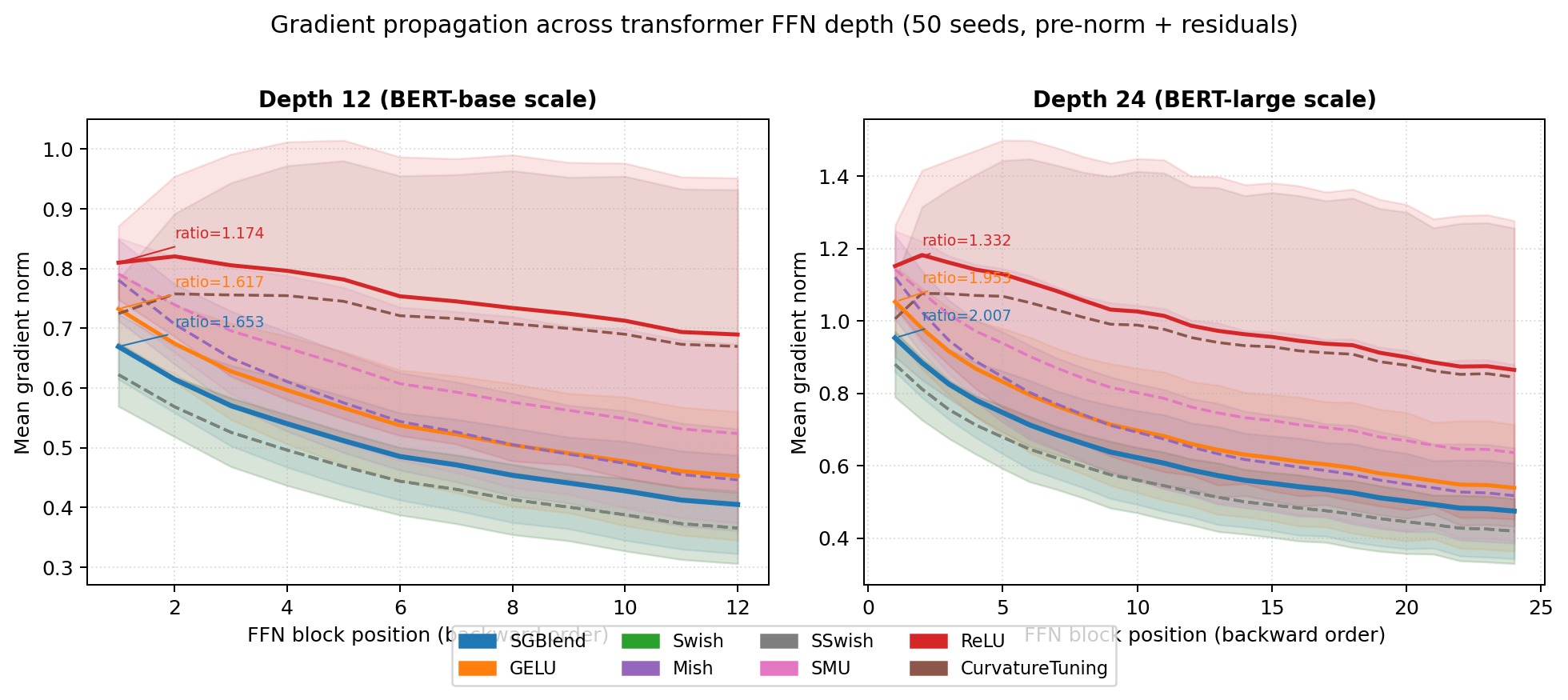}
  \caption{%
    Mean gradient norm per FFN block position at depths 12 and 24,
    estimated over 50~seeds.
    At 24-block depth (right), SG-Blend ($2.007$) sits between GELU
    ($1.953$) and Swish ($2.099$), confirming the blend stays within
    the gradient regime of its components.
    ReLU's lower ratio reflects its well-documented BatchNorm synergy;
    transformer FFN blocks use LayerNorm instead.
  }
  \label{fig:gradient_depth}
\end{figure}

\paragraph{Summary of contributions.}
\begin{enumerate}[noitemsep]
  \item \textbf{SSwish}: a bias-corrected Swish variant with learnable
    sharpness~$\beta$ and bias~$\gamma$ that promotes zero-centered
    activation statistics per layer.
  \item \textbf{SG-Blend}: a per-layer convex interpolation between SSwish
    and GELU with learnable $\alpha$, trained end-to-end; adds only
    3~scalar parameters per layer.
\end{enumerate}


\section{Related Work}
\label{sec:related}

\paragraph{Fixed activation functions.}
ReLU~\citep{nair-2010} remains the dominant choice in BatchNorm-heavy
architectures due to decades of architectural co-optimization.
Swish~\citep{ramachandran-2018} and GELU~\citep{GELU_Hendrycks2016GaussianEL}
are the standard choices in transformer FFN blocks.
Leaky ReLU~\citep{leakyRELU_Maas2013RectifierNI} addresses dead-neuron
pathologies at the cost of a fixed negative slope.
Mish~\citep{misra2019mish} is a self-regularized non-monotonic activation
that achieves strong results on vision tasks but higher per-activation latency.
\citet{DUBEY202292} provide a comprehensive survey and benchmark.

\paragraph{Learnable activation functions.}
PReLU~\citep{PRELU} learns a per-channel negative slope, demonstrating that
even a single learnable scalar per channel can improve ImageNet performance.
\citet{alexandridis2024adaptive} propose Parametric Activation Units (PAU),
learning activation shapes via rational functions.
ParametricSwish, a variant of Swish with a learnable sharpness parameter,
serves as a single-parameter learnable baseline in our experiments.

\paragraph{Curvature Tuning.}
\citet{hu2025curvature} introduced a model steering technique
based on interpolating activation curvature via a single scalar parameter~$\beta$.
Like SG-Blend, they observe that the sharpness of the gating function affects
gradient flow, but their method operates post-training.
SG-Blend differs by training $\beta$ end-to-end alongside $\alpha$ and $\gamma$,
and by providing a per-layer blend between two well-understood activation
families rather than a global curvature shift.

\paragraph{GLU variants and interpolation.}
\citet{shazeer2020glu} systematically explores gated linear unit variants
in transformer FFNs (SwiGLU, GeGLU, etc.), finding that gated formulations
generally outperform their non-gated equivalents.
SG-Blend differs: it operates at the activation level within a standard
two-linear-layer FFN, rather than replacing the FFN structure with a GLU.

\paragraph{Activation adaptation across layers.}
The observation that different layers may benefit from different activation
shapes is implicit in depth-dependent initialization schemes~\citep{LeCun2012}
and the depth gradient we observe in learned $\gamma$ values
(Section~\ref{sec:bert_imdb}).
To our knowledge, SG-Blend is the first activation to explicitly parameterize
a per-layer convex combination of two named activation families with an
interpretable blend coefficient.

\section{Method}
\label{sec:method}

\subsection{SSwish: Bias-Corrected Swish}
\label{sec:sswish}

Swish~\citep{ramachandran-2018} is defined as
$\mathrm{Swish}(x) = x \cdot \sigma(\beta x)$,
where $\sigma$ denotes the sigmoid function and $\beta$ is a sharpness
parameter (fixed at~$1$ in the original formulation).
Swish's mean output is positive, a DC bias that BatchNorm suppresses
automatically but LayerNorm does not, allowing progressive mean shift
across depth~\citep{LeCun2012}.

We define \textbf{SSwish} as:
\begin{equation}
  \mathrm{SSwish}(x;\,\beta,\gamma) = x \cdot \sigma(\beta x) - \gamma,
  \label{eq:sswish}
\end{equation}
where $\beta \in [0.1, 10]$ is a learnable sharpness parameter
(initialized to~$1.0$) and $\gamma \in \mathbb{R}$ is a learnable bias
(initialized to~$0.0$).
The optimizer adjusts $\gamma$ to maintain near-zero mean activation
statistics per layer which is a property LayerNorm alone does not guarantee.
Note that $\gamma$ shifts the output mean, not the functional symmetry
axis; we therefore term this formulation ``bias-corrected Swish.''

\begin{figure}[t]
    \centering
    \includegraphics[width=0.95\linewidth]{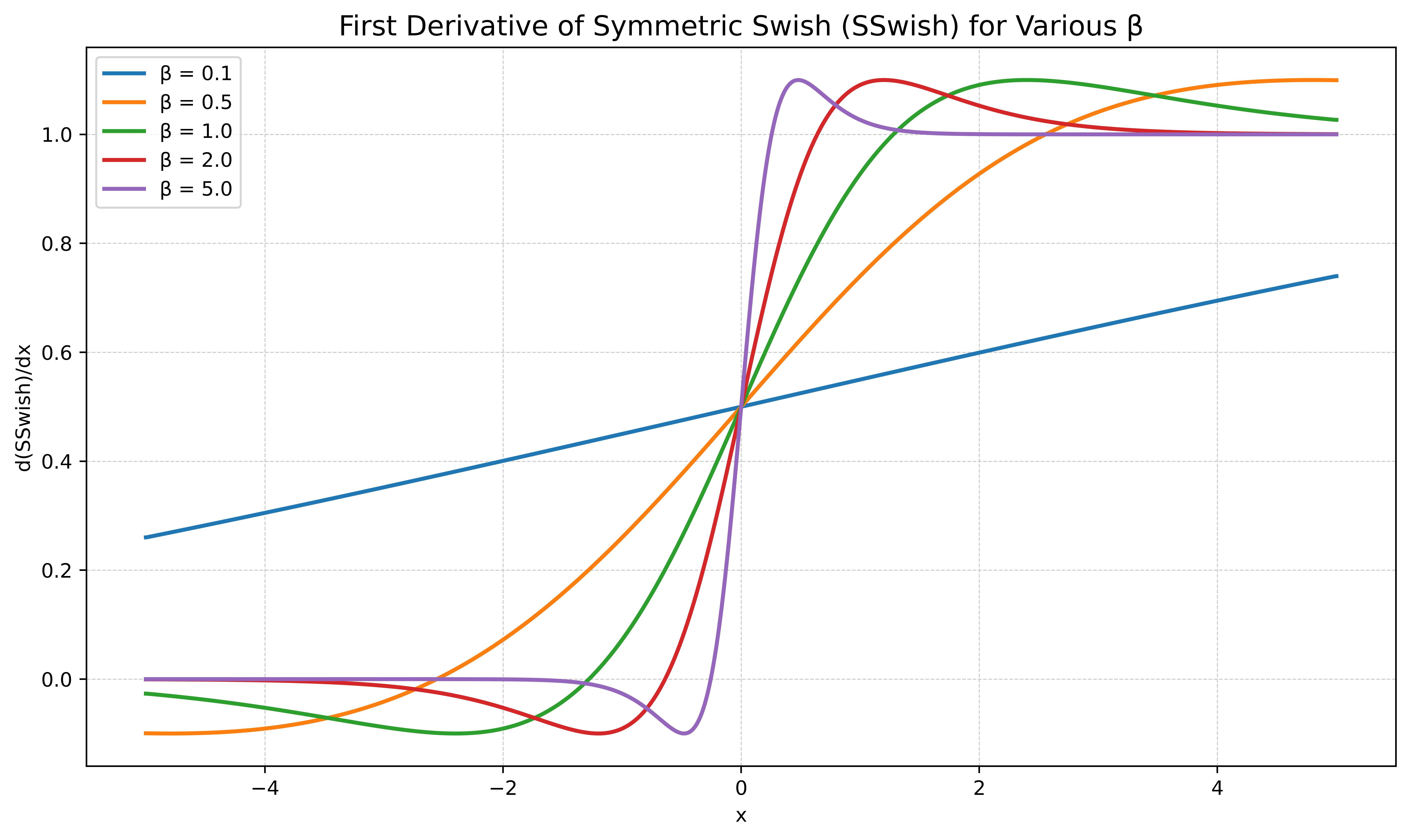} 
    \caption{First derivative of Symmetric Swish (SSwish) for various $\beta$ values (assuming $\gamma=0$). Larger $\beta$ leads to a sharper peak near the origin, influencing gradient flow.}
    \label{fig:sswish_deriv_viz}
\end{figure}

\paragraph{First derivative.}
\begin{equation}
  \frac{d}{dx}\mathrm{SSwish}(x) =
    \sigma(\beta x) + \beta x\,\sigma(\beta x)(1 - \sigma(\beta x)),
\end{equation}
which is bounded and non-negative for $x \geq 0$, as shown in Fig. \ref{fig:sswish_deriv_viz} confirming that SSwish
avoids the dead-gradient problem of ReLU for positive inputs while remaining
differentiable everywhere. SSwish is infinitely differentiable ($C^\infty$) due to the smoothness of the sigmoid function, which is beneficial for gradient-based optimization.

\begin{figure}[t]
  \centering
  \includegraphics[width=\columnwidth]{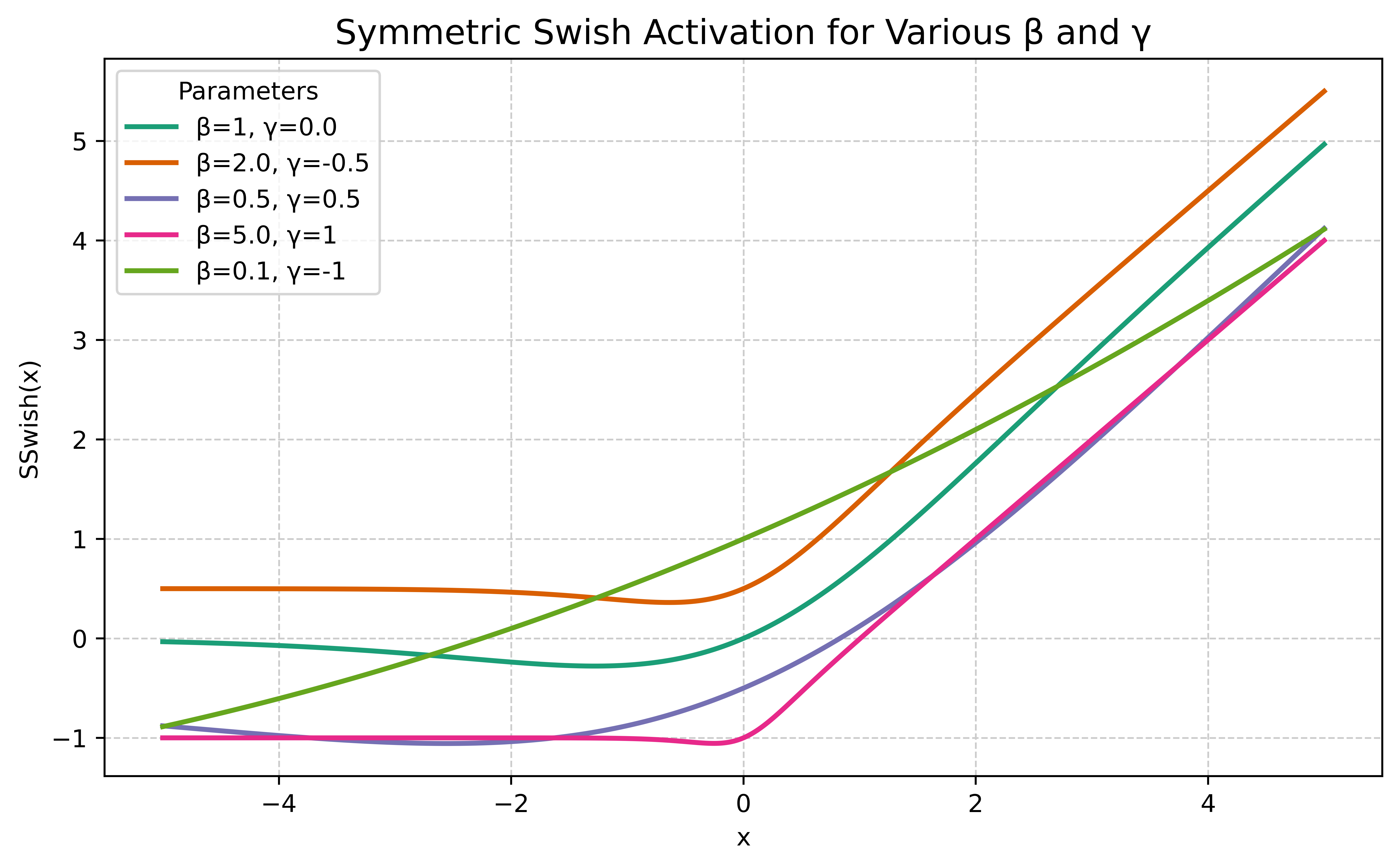}
  \caption{SSwish output for varying $\beta$ and $\gamma$.
    Higher $\beta$ sharpens the gate; positive $\gamma$ shifts the output
    downward, promoting zero-centered activation statistics.}
  \label{fig:sswish_shape}
\end{figure}

\begin{table}[t]
\centering
\begin{tabularx}{\columnwidth}{@{}l XX@{}}
\toprule
\textbf{Parameter} & \textbf{Effect} & \textbf{Derivative Impact} \\
\midrule
$\beta \uparrow$   & Sharper transition & Larger peak gradient \\
$\beta \downarrow$ & Flatter, smoother   & Smaller derivative \\
$\gamma \uparrow$  & Downward shift      & No effect on derivative \\
$\gamma \downarrow$ & Upward shift        & No effect on derivative \\
\bottomrule
\end{tabularx}
\caption{Effect of Learnable Parameters on Symmetric Swish}
\label{tab:symmetric-swish}
\end{table}

\subsection{SG-Blend: Convex Interpolation}
\label{sec:sgblend}

\textbf{SG-Blend} interpolates between SSwish and GELU:
\begin{equation}
  \begin{split}
    \mathrm{SGBlend}(x;\,\alpha,\beta,\gamma) = {} & \alpha \cdot \mathrm{SSwish}(x;\,\beta,\gamma) \\
    & + (1-\alpha) \cdot \mathrm{GELU}(x),
  \end{split}
  \label{eq:sgblend}
\end{equation}
where $\alpha \in [0,1]$ is a per-layer learnable blend coefficient
(initialized to~$0.5$).
Each transformer FFN block maintains its own independent copy of
$(\alpha, \beta, \gamma)$, adding only \textbf{3~scalar parameters} per block.

\begin{table}[h]
\centering
\begin{tabularx}{\columnwidth}{@{}l XX@{}}
\toprule
\textbf{Parameter} & \textbf{Effect} & \textbf{Derivative Impact} \\
\midrule
$\alpha = 1$       & Pure SSwish & Non-monotonic, sharp gradients \\
$\alpha = 0$       & Pure GELU & Smoother transition, less peaky \\
$0 < \alpha < 1$   & Smooth interpolation & Balanced non-linearity \\
$\beta \uparrow$   & Sharper SSwish slope & Stronger gradient peaks \\
$\beta \downarrow$ & Softer, sigmoid-like curve & Lower derivative \\
$\gamma \uparrow$  & Downward output shift & No gradient effect \\
$\gamma \downarrow$& Upward output shift & No gradient effect \\
\bottomrule
\end{tabularx}
\caption{Effect of SG-Blend Parameters}
\label{tab:sg-blend-parameters}
\end{table}

\paragraph{Interpretation.}
When $\alpha{=}0$, SG-Blend reduces to pure GELU; when $\alpha{=}1$, it
reduces to SSwish.
Intermediate $\alpha$ lets each layer locate its own optimum in the
SSwish--GELU activation space.
Because both endpoints are well-understood activations, $\alpha$ is directly
interpretable as a \emph{per-layer activation-family preference score}.
Section~\ref{sec:param_analysis} shows that layers can converge to
$\alpha \approx 0.5$ (equal blend), confirming that neither pure endpoint
is uniformly optimal.

\paragraph{First Derivative.}
Using the chain rule and noting that $\mathrm{GELU}' = \Phi(x) + x\phi(x)$
(where $\Phi$ and $\phi$ are the standard normal CDF and PDF):
\begin{align}
  \frac{d}{dx}\mathrm{SGBlend}(x) = {} & \alpha \cdot \frac{d}{dx}\mathrm{SSwish}(x) \nonumber \\
  & + (1-\alpha)\bigl[\Phi(x) + x\phi(x)\bigr].
  \label{eq:sgblend_grad}
\end{align}
Both summands are bounded and continuous, so SG-Blend inherits the
smooth gradient properties of its components.

\begin{figure}[t]
  \centering
  \includegraphics[width=\columnwidth]{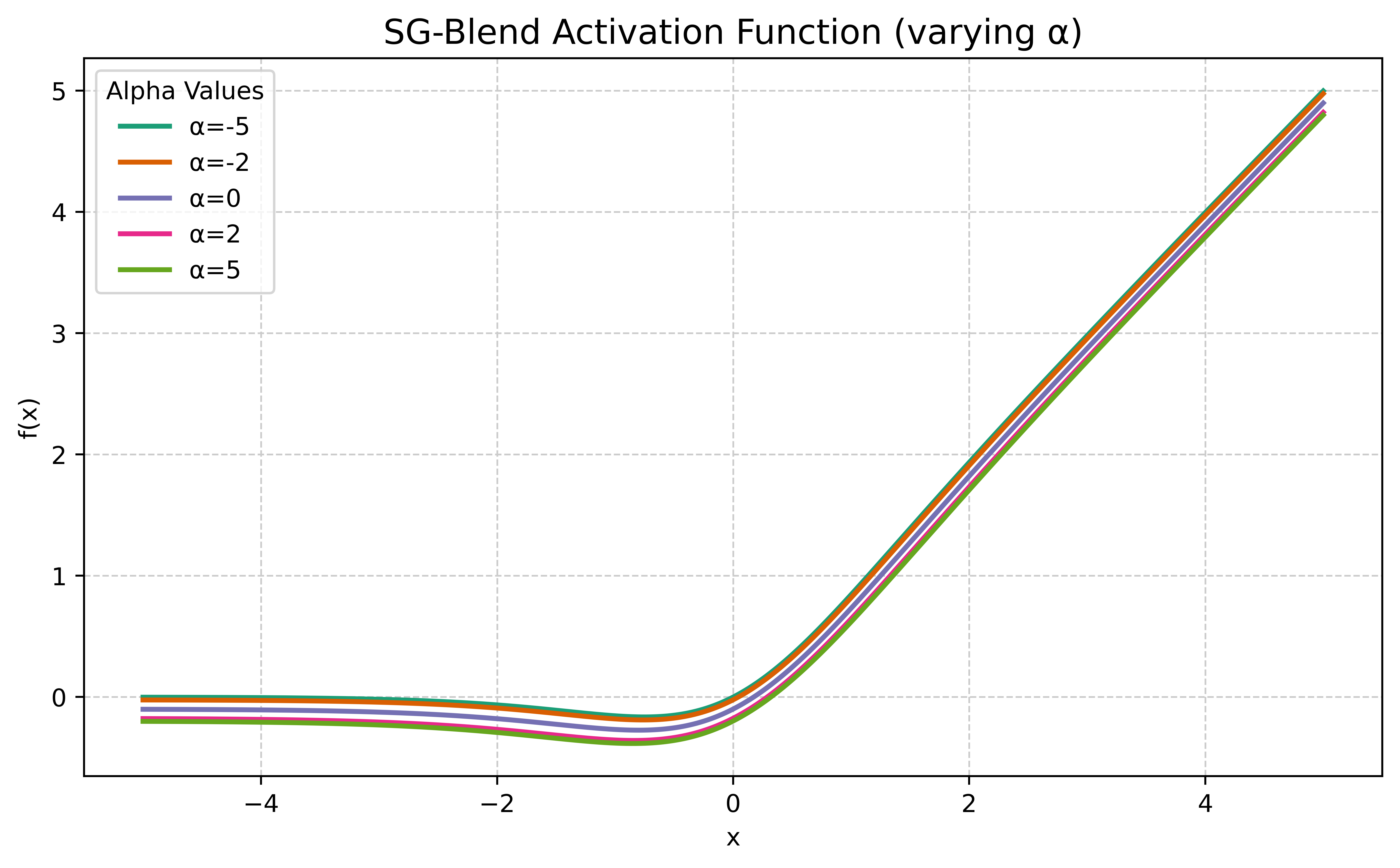}
  \caption{SG-Blend output for varying $\alpha$ with $\beta{=}1$,
    $\gamma{=}0$.
    $\alpha{=}0$ recovers GELU; $\alpha{=}1$ recovers SSwish.
    Intermediate $\alpha$ produces smooth interpolants.}
  \label{fig:sgblend_alpha}
\end{figure}

\begin{figure}[t]
    \centering
    \includegraphics[width=\columnwidth]{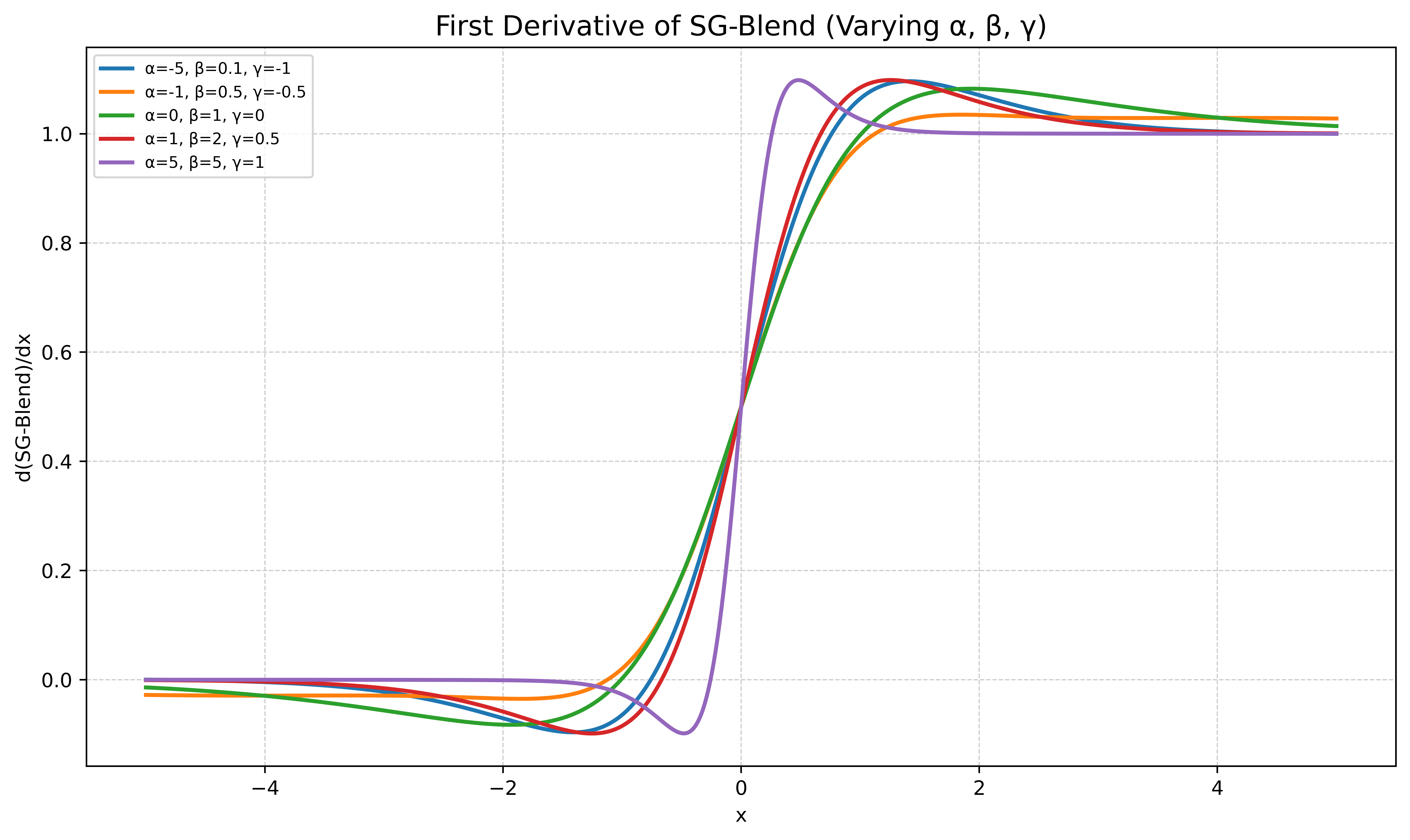} 
    \caption{First derivative of SG-Blend activation function shape for varying $\alpha$, $\beta$ and $\gamma$.}
    \label{fig:sgblend_params_viz}
\end{figure}


\subsection{Implementation Details}
\label{sec:implementation}

SG-Blend is implemented in PyTorch~\citep{pytorch-paszke-2019}.
The three parameters are learned end-to-end: $\alpha$ is constrained to
$(0,1)$ via a sigmoid reparameterization; $\beta$ is projected to
$[0.1,10]$ after each gradient step.
Swapping SG-Blend into any deep learning module requires replacing the
activation call with an instantiated \texttt{SGBlend} module; no
architectural changes are needed.

\paragraph{GELU approximation.}
The GELU component inside SG-Blend uses the sigmoid approximation
$x\,\sigma(1.702\,x)$~\citep{GELU_Hendrycks2016GaussianEL} rather than the exact
erf-based formulation; the non-learnable GELU baseline uses the exact
implementation.
The approximation is differentiable everywhere with a closed-form
gradient, which simplifies joint optimization with $\alpha$ and $\beta$ without introducing any meaningful degradation in accuracy.
Reported GELU baselines are computed with the exact implementation
throughout.

\begin{table}[t]
  \centering
  \small
  \begin{tabular}{@{}lccc@{}}
    \toprule
    \textbf{Parameter} & \textbf{Init} & \textbf{Constraint} & \textbf{Role} \\
    \midrule
    $\alpha$ & $0.5$ & $(0,1)$ via sigmoid & blend coefficient \\
    $\beta$  & $1.0$ & $[0.1,10]$ & gate sharpness \\
    $\gamma$ & $0.0$ & $\mathbb{R}$ & output bias shift \\
    \bottomrule
  \end{tabular}
  \caption{SG-Blend learnable parameters.}
  \label{tab:params}
\end{table}

\section{Experiments}
\label{sec:experiments}

We evaluate SG-Blend through seven experiments designed to highlight its salient features in diverse setting.

\subsection{Experimental Setup}

\paragraph{Baselines.}
We compare against: \textbf{ReLU}~\citep{nair-2010},
\textbf{GELU}~\citep{GELU_Hendrycks2016GaussianEL},
\textbf{Swish}~\citep{ramachandran-2018},
\textbf{Mish}~\citep{misra2019mish},
\textbf{SSwish} (ours, Section~\ref{sec:sswish}),
\textbf{ParametricSwish} (learnable-$\beta$ Swish),
\textbf{PReLU}~\citep{PRELU},
and \textbf{CurvatureTuning}~\citep{hu2025curvature}.
This covers monotonic, non-monotonic, fixed-shape, and multi-parameter
learnable activations.
\paragraph{Hardware.}
All experiments ran on a single NVIDIA T4 GPU (16\,GB); the gradient
stability study is CPU-feasible and was run on CPU.

\subsection{Gradient Stability in Deep Transformer FFN Stacks}
\label{sec:gradient_stability}

\paragraph{Setup.}
We construct transformer FFN stacks of depth~12 and depth~24 (matching
BERT-base and BERT-large block counts) with pre-LayerNorm and residual
connections copying the standard transformer FFN block structure.
We measure the mean gradient norm at each block position after one backward
pass and report the \emph{gradient propagation ratio}
(first-block norm $\div$ last-block norm): a ratio closer to~1 indicates
more uniform gradient flow; larger ratios indicate more differential scaling
between early and late layers.

\paragraph{Results.}
Table~\ref{tab:gradient} shows the gradient propagation ratios.
Figure~\ref{fig:gradient_depth} visualizes the per-block gradient norm curves.

\begin{table}[t]
  \centering
  \small
  \setlength{\tabcolsep}{3pt}
  \begin{tabular}{@{}lcccc@{}}
    \toprule
    & \multicolumn{2}{c}{\textbf{Depth 12}} & \multicolumn{2}{c}{\textbf{Depth 24}} \\
    \cmidrule(lr){2-3} \cmidrule(l){4-5}
    \textbf{Activation} & \textbf{Ratio} & \textbf{Final Std} & \textbf{Ratio} & \textbf{Final Std} \\
    \midrule
    \textbf{SG-Blend}  & $1.653$ & $0.082$ & $2.007$ & $0.132$ \\
    GELU               & $1.617$ & $0.108$ & $1.953$ & $0.175$ \\
    Swish              & $1.702$ & $0.060$ & $2.099$ & $0.090$ \\
    Mish               & $1.749$ & $0.085$ & $2.166$ & $0.132$ \\
    SSwish             & $1.702$ & $0.060$ & $2.099$ & $0.090$ \\
    SMU                & $1.510$ & $0.150$ & $1.796$ & $0.243$ \\
    CurvatureTuning    & $1.082$ & $0.262$ & $1.191$ & $0.412$ \\
    ReLU               & $1.174$ & $0.262$ & $1.332$ & $0.412$ \\
    \bottomrule
  \end{tabular}
  \caption{%
    Gradient propagation ratio (first-block norm $\div$ last-block norm) and the final-block standard deviation, estimated over 50~seeds.
    While ReLU and CurvatureTuning exhibit ratios closer to $1.0$, they suffer from explosive cross-seed variance in this Pre-LayerNorm setting (final-block std near $0.41$). Among stable activations, SG-Blend ($2.007$) and GELU ($1.953$) achieve the highest uniformity at depth~24. SG-Blend cleanly interpolates between GELU and Swish/SSwish.
  }
  \label{tab:gradient}
\end{table}

In standard CNNs with BatchNorm, a gradient propagation ratio of $1.0$ is ideal. However, in Pre-LayerNorm architectures, gradients naturally decay as they pass backward through successive normalization layers in the residual branch, making a ratio of $1.0$ structurally unnatural. While ReLU ($1.332$) and CurvatureTuning ($1.191$) numerically approach $1.0$, our cross-seed data reveals this comes at the cost of catastrophic variance: both exhibit standard deviations approaching $50\%$ of their mean gradient norms at the final block. By fighting the natural gradient decay of the LayerNorm architecture, they render the network highly sensitive to random initialization.

Among the activations that maintain cross-seed mathematical stability, GELU ($1.953$) and our proposed SG-Blend ($2.007$) demonstrate the highest functional uniformity at depth 24. SG-Blend maintains substantially more stable gradient flows than Mish ($2.166$), Swish ($2.099$), and SSwish ($2.099$). Its gradient behavior cleanly \emph{interpolates} between GELU and Swish, confirming it does not degrade gradient characteristics beyond either pure component.

\subsection{BERT Sentiment Analysis on IMDB}
\label{sec:bert_imdb}

\begin{figure}[t]
  \centering
  \includegraphics[width=\columnwidth]{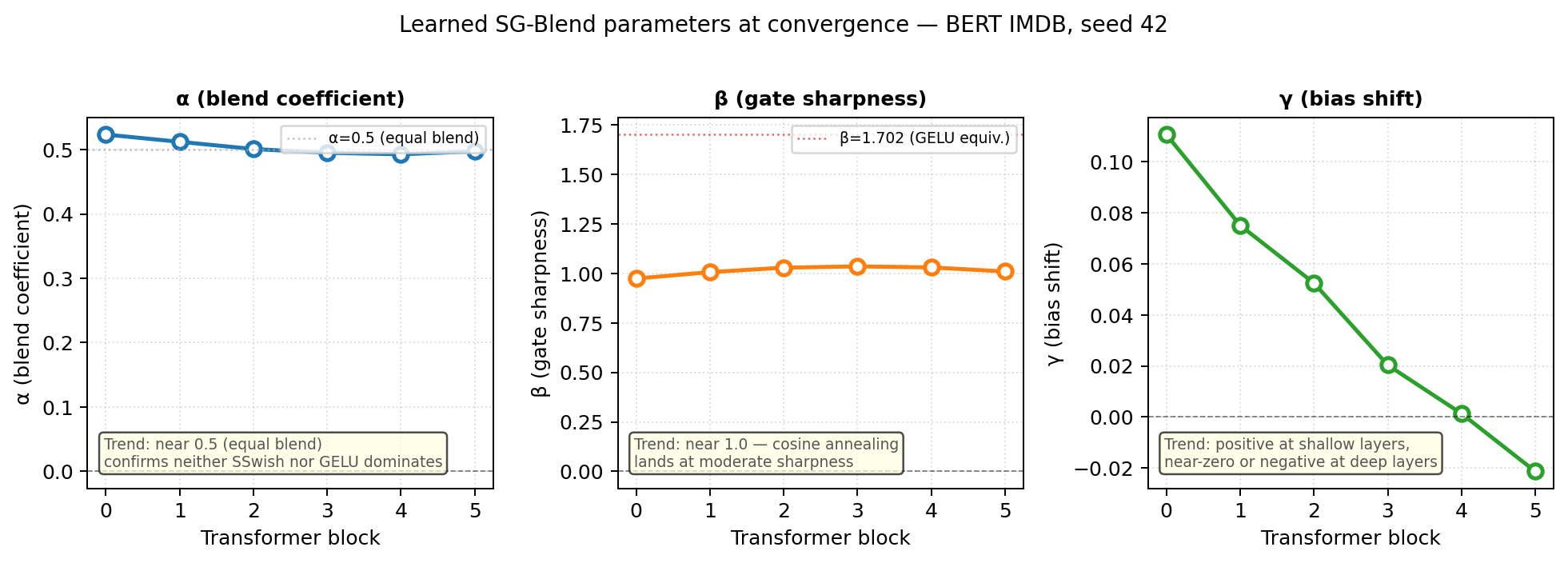}
  \caption{%
    Learned $\alpha$, $\beta$, $\gamma$ at convergence for seed~42
    (BERT IMDB).
    $\alpha \approx 0.5$ across all layers (equal blend);
    $\gamma$ shows a depth gradient (positive $\to$ near-zero), indicating
    that early layers benefit more from output centering than deep layers.
  }
  \label{fig:param_evolution}
\end{figure}

\paragraph{Setup.}
We train a BERT-style transformer (6~encoder blocks, $d_\text{model}{=}128$,
4~heads, 2-layer FFN per block) from scratch on the IMDB sentiment dataset
\citep{imdb-maas2011} (25K training examples, 25k testing examples).
Experiments use the Adam optimizer~\citep{adam-Kingma2014AdamAM} with early
stopping (patience~5).
We run \textbf{5 seeds} per activation (seeds 42, 123, 456, 789, 1024) and
report mean $\pm$ standard deviation of test accuracy.
\paragraph{Results.}
Table~\ref{tab:bert_imdb} shows the metrics, and the corresponding architectural dynamics are illustrated in Figure~\ref{fig:param_evolution}.

\begin{table}[t]
  \centering
  \small
  \begin{tabular}{@{}lcc@{}}
    \toprule
    \textbf{Activation} & \textbf{Mean (\%)} & \textbf{Std (pp)} \\
    \midrule
    Swish              & \textbf{81.32} & 0.37          \\
    \textbf{SG-Blend}  & \textbf{81.31} & \textbf{0.27} \\
    GELU               & 81.23          & 0.47          \\
    CurvatureTuning    & 81.17          & 0.27          \\
    SMU                & 81.07          & 0.55          \\
    ReLU               & 81.05          & 0.38          \\
    Mish               & 81.05          & 0.55          \\
    ParametricSwish    & 81.00          & 0.90          \\
    SSwish             & 80.91          & 0.09          \\
    PReLU              & 80.60          & 0.82          \\
    \bottomrule
  \end{tabular}
  \caption{%
    BERT IMDB test accuracy (\%), 5~seeds per activation.
    SG-Blend matches peak accuracy while achieving the lowest variance
    among competitive activations (tied with CurvatureTuning at ${\pm}0.27$\,pp)---a
    meaningful advantage for practitioners who run a single seed.
  }
  \label{tab:bert_imdb}
\end{table}

\paragraph{Accuracy.}
SG-Blend is statistically tied with Swish at the top of the ranking
($81.31\%$ vs.\ $81.32\%$, within one standard deviation of each other).
SG-Blend outperforms GELU by $0.08$\,pp and Swish outperforms GELU by $0.09$\,pp on
average; neither difference is individually significant over 5~seeds.

\paragraph{Variance.}
SG-Blend's $\pm0.27$\,pp standard deviation is $\mathbf{42\%}$ lower than
GELU's $\pm0.47$\,pp at comparable mean accuracy.
The gap is largest at the worst case: a practitioner who runs one seed
would see $80.81\%$ with SG-Blend versus $80.46\%$ with GELU: a
$0.35$\,pp difference large enough to reverse a head-to-head comparison.

\paragraph{Learned parameter analysis.}
\label{sec:param_analysis}
Figure~\ref{fig:param_evolution} shows the learned $\alpha$, $\beta$, and
$\gamma$ values at convergence for seed~42.
All layers settle near $\alpha \approx 0.5$, confirming that neither pure
SSwish nor pure GELU dominates, the interpolation finds an intermediate
optimum.
$\beta$ converges near~$1.0$ across all layers.
$\gamma$ shows a clear depth gradient: positive (${\sim}0.09$) at the
earliest block, decreasing monotonically to near-zero
(${\sim}{-}0.03$) at block~5.
This suggests that early-layer activations benefit more from a downward bias
shift than deep-layer activations: a layer-specific adaptation that
SG-Blend's architecture enables but fixed activations cannot.

\subsection{GPT-2-Style Language Model Pretraining on WikiText-103}
\label{sec:lm_pretrain}

To assess SG-Blend at LLM-scale pretraining, we train a GPT-2-style
decoder-only transformer from scratch and compare validation perplexity
across activation functions on WikiText-103~\citep{wikitext}.

\paragraph{Setup.}
We use the \emph{tiny} model configuration: 6 layers, 6 attention heads,
embedding dimension 384, totalling ${\approx}25$M parameters (comparable
in scale to a compact GPT-2).
Text is tokenized with the GPT-2 BPE tokenizer (vocabulary size 50{,}304).
Training runs for 100{,}000 steps with batch size~8 and sequence length~512,
consuming ${\approx}400$M tokens.
We use AdamW ($\beta_1{=}0.9$, $\beta_2{=}0.95$, weight decay~$0.1$) with a
cosine learning-rate schedule (peak $6\times10^{-4}$, 2{,}000 warmup steps,
minimum $6\times10^{-5}$) and mixed-precision (FP16) training on a T4 GPU.
A single seed is run per activation (seed~42).

\begin{table}[t]
  \centering
  \small
  \begin{tabular}{@{}lcc@{}}
    \toprule
    \textbf{Activation} & \textbf{Val.\ Loss} & \textbf{Val.\ PPL} \\
    \midrule
    \textbf{SG-Blend}  & \textbf{3.894}   & \textbf{49.10} \\
    GELU               & 3.901               & 49.47             \\
    ParametricSwish    & 3.908               & 49.82             \\
    SSwish             & 3.909               & 49.87             \\
    Swish              & 3.958               & 52.33             \\
    \bottomrule
  \end{tabular}
  \caption{%
    Validation perplexity on WikiText-103 after 100K training steps
    with a ${\approx}25$M-parameter GPT-2-style model (single seed).
    Lower is better. \textbf{Bold}: best;
  }
  \label{tab:lm_pretrain}
\end{table}

SG-Blend achieves the lowest perplexity ($49.10$), outperforming
GELU ($49.47$), ParametricSwish ($49.82$), SSwish ($49.87$), and Swish
($52.33$).
The $0.37$-point perplexity advantage of SG-Blend over GELU is consistent
with the gradient-stability findings in Section~\ref{sec:gradient_stability}:
better-conditioned activations sustain effective learning over long token
horizons even in a small model.
These results indicate that SG-Blend's advantage is not limited to
fine-tuning or classification tasks but extends to the autoregressive
pretraining regime.

\subsection{Ablation Study}
\label{sec:ablation}

\paragraph{Setup.}
We isolate the contribution of each learnable parameter using a 2-block
Transformer encoder on IMDB (same dataset, single seed, fast iteration).
We compare: full SG-Blend, fixed $\alpha{=}0.5$ (blend structure without
learned blending), fixed $\beta{=}1.702$ (GELU-equivalent sharpness),
fixed $\gamma{=}0$ (no bias shift), pure GELU ($\alpha \to 0$), pure SSwish
($\alpha \to 1$), per-neuron variant, and a non-learnable GELU baseline.

\paragraph{Results.}

\begin{table}[t]
  \centering
  \small
  \begin{tabular}{@{}lcc@{}}
    \toprule
    \textbf{Configuration} & \textbf{Acc.\ (\%)} & \textbf{$\Delta$} \\
    \midrule
    \textbf{SG-Blend full} ($\alpha,\beta,\gamma$ learned) & \textbf{77.09} & $+0.11$ \\
    Fixed $\alpha{=}0.5$, $\beta$+$\gamma$ learned         & 77.04          & $+0.06$ \\
    Fixed $\beta{=}1.702$, $\alpha$+$\gamma$ learned       & 76.97          & $-0.01$ \\
    Per-neuron SG-Blend                                     & 76.91          & $-0.07$ \\
    GELU baseline (non-learnable)                           & 76.98          & $\pm0.00$ \\
    Fixed $\gamma{=}0$, $\alpha$+$\beta$ learned           & 76.78          & $-0.20$ \\
    Pure GELU ($\alpha \to 0$ fixed)                       & 76.90          & $-0.08$ \\
    Pure SSwish ($\alpha \to 1$ fixed)                     & 76.71          & $-0.27$ \\
    \bottomrule
  \end{tabular}
  \caption{%
    Ablation of SG-Blend components.
    Accuracy on IMDB (single seed).
    $\Delta$ = difference from GELU baseline.
  }
  \label{tab:ablation}
\end{table}

\paragraph{Key findings.}
(1)~\textbf{The interpolation structure is the core contribution}: even with
$\alpha$ frozen at $0.5$, the blend outperforms both pure GELU and pure
SSwish ($+0.06$\,pp vs.\ $-0.08$\,pp and $-0.27$\,pp respectively),
confirming that the architectural blend, not just learned $\alpha$, provides
value.
(2)~Fixing $\beta{=}1.702$ (the GELU-equivalent value) nearly matches the
GELU baseline, confirming that $\beta$ needs freedom to deviate from $1.702$
for the SSwish component to contribute.

\begin{figure}[t]
  \centering
  \includegraphics[width=\columnwidth]{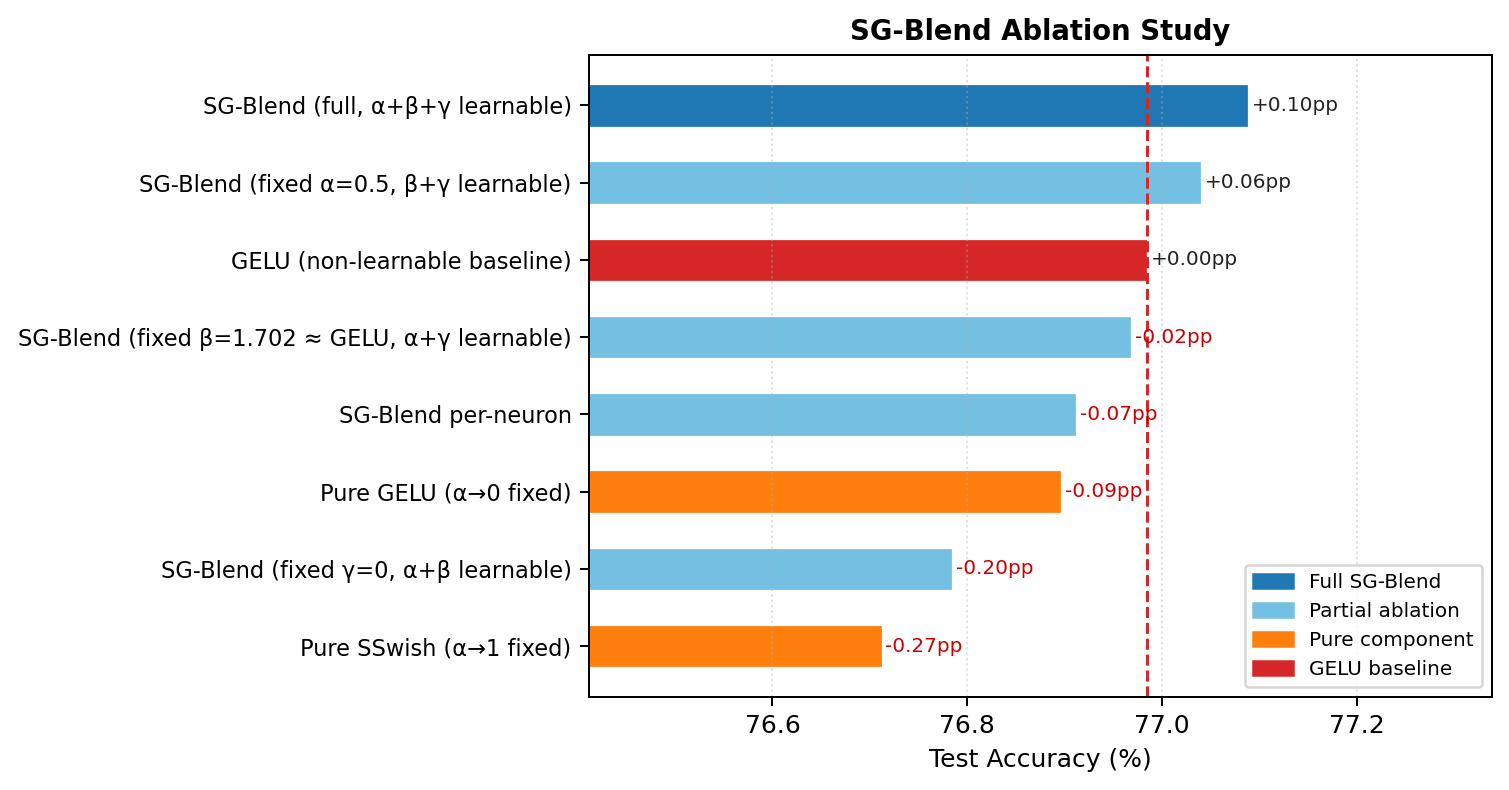}
  \caption{Ablation of SG-Blend parameters.
    Even with $\alpha$ fixed at~$0.5$, the blend outperforms either
    pure component, isolating the interpolation structure as the contribution.}
  \label{fig:ablation}
\end{figure}

\subsection{Image Classification Performance}
\label{sec:vision_results}

SG-Blend's performance was evaluated on CIFAR-10 using ResNet18 and ResNet50. Table~\ref{tab:cifar_results} presents the test accuracies.

\begin{table}[h]
\centering
\begin{tabular}{@{}lcc@{}}
\toprule
\textbf{Activation} & \textbf{ResNet18 (\%)} & \textbf{ResNet50 (\%)} \\
\midrule
\textbf{SGBlend}    & \textbf{93.23}         & \textbf{92.71}         \\
ReLU                & 92.79                  & 92.03                  \\
Swish               & 92.87                  & 91.58                  \\
GELU                & 92.02                  & 91.39                  \\
Mish                & 92.26                  & 91.90                  \\
\bottomrule
\end{tabular}
\caption{Test Accuracy (\%) on CIFAR-10 with ResNet Architectures. Best results for each configuration are in \textbf{bold}.}
\label{tab:cifar_results}
\end{table}

On CIFAR-10, SG-Blend (93.23\%) consistently achieves the highest test accuracy. With ResNet18, surpasses the next best baseline, Swish (92.87\%). Using the deeper ResNet50 architecture, SG-Blend (92.71\%) again outperforms all baselines, including ReLU (92.03\%) and Mish (91.90\%).

\subsection{CIFAR-100 with WideResNet-28-10}
\label{sec:cifar100}

To assess SG-Blend outside the transformer domain, we run a single-seed
WideResNet-28-10~\citep{zagoruyko2016wide} experiment on CIFAR-100
(200~epochs, SGD with momentum, cosine annealing LR).

\begin{table}[t]
  \centering
  \small
  \begin{tabular}{@{}lc@{}}
    \toprule
    \textbf{Activation} & \textbf{Best Acc.\ (\%)} \\
    \midrule
    ReLU               & \textbf{80.36} \\
    SSwish             & 77.46          \\
    \textbf{SG-Blend}  & 77.33          \\
    SMU                & 77.13          \\
    GELU               & 77.01          \\
    ParametricSwish    & 76.78          \\
    PReLU              & 75.95          \\
    Mish               & 75.73          \\
    Swish              & 75.07          \\
    \bottomrule
  \end{tabular}
  \caption{%
    WideResNet-28-10 best test accuracy on CIFAR-100.
    ReLU retains its lead in the CNN+BatchNorm setting.
  }
  \label{tab:cifar100}
\end{table}

ReLU wins by a large margin ($80.36\%$ vs.\ SG-Blend's $77.33\%$).
This is expected: CNNs trained with BatchNorm have decades of hyperparameter
tuning and architectural co-optimization for ReLU.
BatchNorm normalizes pre-activations across the batch, partially mitigating
the gradient-stability differences between activation functions that SG-Blend
is designed to address.
In transformer FFN blocks, which use LayerNorm rather than BatchNorm,
this co-optimization is absent,  which is precisely the setting for which
SG-Blend is targeted.
SG-Blend beats GELU ($77.33\%$ vs.\ $77.01\%$), SMU ($77.13\%$), and Mish
($75.73\%$) in this single-seed run, but we caution against drawing strong
conclusions from a single seed.

\subsection{Computational Overhead}
\label{sec:overhead}

\paragraph{Setup.}
We benchmark all activations on a tensor of shape $(256, 128, 256)$
(transformer-shaped batch), with 100 warmup and 500 timed repetitions on
the same T4 GPU.

\begin{table}[t]
  \centering
  \small
  \begin{tabular}{@{}lccc@{}}
    \toprule
    \textbf{Activation} & \textbf{Fwd (ms)} & \textbf{Bwd (ms)} & \textbf{Train (s/seed)} \\
    \midrule
    GELU               & 0.012 & 0.070 & ${\approx}45$ \\
    Swish              & 0.012 & 0.069 & ${\approx}45$ \\
    ReLU               & 0.012 & 0.069 & ${\approx}45$ \\
    Mish               & 0.059 & 0.170 & ${\approx}55$ \\
    SSwish             & 0.055 & 0.242 & ${\approx}52$ \\
    SMU                & 0.107 & 0.458 & ${\approx}58$ \\
    \textbf{SG-Blend}  & \textbf{0.143} & \textbf{0.532} & ${\approx}60$ \\
    \bottomrule
  \end{tabular}
  \caption{%
    Per-activation latency (ms) and end-to-end BERT IMDB training time.
    SG-Blend adds ${\approx}33\%$ end-to-end overhead vs.\ GELU.
  }
  \label{tab:overhead}
\end{table}

Per-activation, SG-Blend is $11.7\times$ GELU in forward and $7.7\times$
in backward latency, because it must evaluate both the SSwish and GELU
forward passes separately.
End-to-end, this overhead amortizes across linear layers, attention, and I/O:
BERT training takes ${\approx}60$\,s/seed with SG-Blend vs.\ ${\approx}45$\,s
with GELU---a $\mathbf{33\%}$ increase.
For comparison, Mish (widely used in practice) adds ${\approx}22\%$
end-to-end, and SMU adds ${\approx}29\%$.
The overhead is therefore comparable in magnitude to other multi-parameter
learnable activations.

\begin{figure}[t]
  \centering
  \includegraphics[width=\columnwidth]{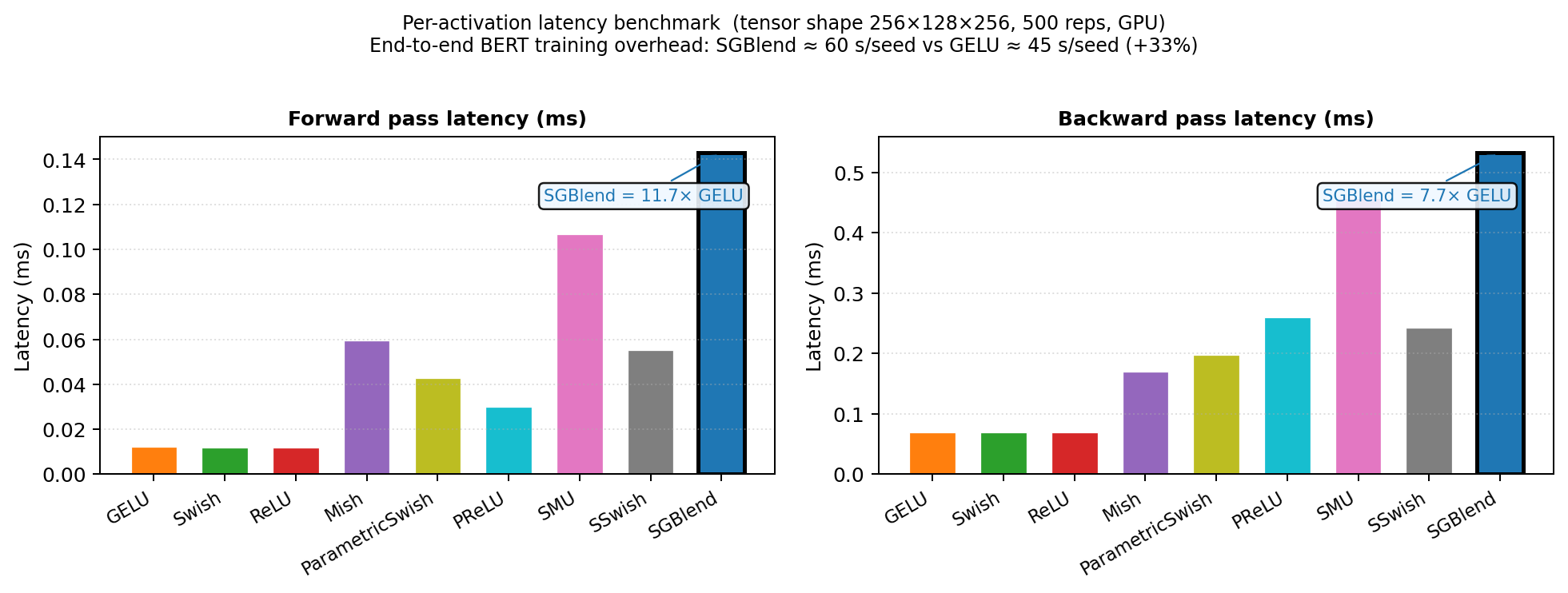}
  \caption{Per-activation forward and backward latency.
    SG-Blend adds ${\approx}33\%$ end-to-end training overhead vs.\ GELU,
    comparable to other multi-parameter learnable activations.}
  \label{fig:overhead}
\end{figure}

\section{Conclusion}
\label{sec:conclusion}

We presented SG-Blend, a per-layer adaptive activation that interpolates
between a bias-corrected Swish variant (SSwish) and GELU via three learnable
scalars per transformer block.
We present \emph{reliable top-tier
performance in various deep learning models.}

The key empirical findings are:
(1)~SG-Blend's gradient propagation ratio ($2.007$ at depth-24) interpolates
between GELU ($1.953$) and Swish ($2.099$), confirming the blend stays within
the gradient regime of its components and does not degrade gradient flow
beyond either pure activation (50~seeds);
(2)~$42\%$ lower seed-to-seed variance than GELU on BERT IMDB at matched
mean accuracy, a practical reproducibility improvement for single-seed
practitioners;
(3)~an ablation showing that even a frozen $\alpha{=}0.5$ blend outperforms
either pure component, isolating the interpolation structure as the contribution;
(4)~SG-Blend generalizes effectively to autoregressive language modeling, achieving the lowest validation perplexity (49.10) in a GPT-2-style WikiText-103 pretraining setup, outperforming pure GELU and Swish;
(5)~learned parameter analysis reveals a clear depth gradient for the bias parameter $\gamma$, proving that early transformer layers benefit from output centering in ways deeper layers do not—an adaptation impossible with fixed-shape activations.




\section*{Limitations}

The main limitations of SG-Blend are as follows.

\textbf{Computational overhead.}
SG-Blend incurs ${\approx}33\%$ end-to-end training time increase versus
GELU on a T4 GPU ($11.7\times$ per-activation forward latency), making it
the primary practical barrier to adoption.
Users must evaluate whether the variance-reduction benefit ($42\%$ lower
seed-to-seed standard deviation on BERT IMDB) justifies this cost for their
specific use case.
We do not recommend SG-Blend in latency-sensitive inference settings without
further engineering (e.g., operator fusion).

\textbf{Vision scope.}
Our WideResNet results are single-seed, and we have not evaluated SG-Blend
on large-scale vision benchmarks; broader conclusions about vision
performance should therefore be drawn with caution.



\bibliography{custom}

\appendix

\section{Learned Parameter Statistics Across Seeds}
\label{sec:param_stats}

Table~\ref{tab:param_stats} reports the mean and standard deviation of
learned $\alpha$, $\beta$, and $\gamma$ across 5~seeds at convergence
for the BERT IMDB model.
All layers converge to $\alpha \approx 0.5$, confirming the equal-blend
optimum is robust across random initializations.
The $\gamma$ depth gradient is consistent across seeds.


\begin{table}[h]
  \centering
  \small
  \begin{tabular}{@{}lccc@{}}
    \toprule
    \textbf{Block} & $\alpha$ & $\beta$ & $\gamma$ \\
    \midrule
    0 & $\phantom{-}0.517 \pm 0.003$ & $0.987 \pm 0.024$ & $\phantom{-}0.091 \pm 0.017$ \\
    1 & $\phantom{-}0.510 \pm 0.003$ & $1.003 \pm 0.016$ & $\phantom{-}0.067 \pm 0.012$ \\
    2 & $\phantom{-}0.502 \pm 0.003$ & $1.037 \pm 0.013$ & $\phantom{-}0.055 \pm 0.025$ \\
    3 & $\phantom{-}0.497 \pm 0.005$ & $1.034 \pm 0.027$ & $\phantom{-}0.032 \pm 0.010$ \\
    4 & $\phantom{-}0.491 \pm 0.003$ & $1.036 \pm 0.013$ & $-0.001 \pm 0.005$ \\
    5 & $\phantom{-}0.498 \pm 0.002$ & $1.007 \pm 0.006$ & $-0.029 \pm 0.013$ \\
    \bottomrule
  \end{tabular}
  \caption{%
    Learned SG-Blend parameters at convergence (mean $\pm$ std across
    5~seeds, BERT IMDB).
  }
  \label{tab:param_stats}
\end{table}


\end{document}